\documentclass[runningheads]{llncs}

\usepackage{comment}
\usepackage{amsmath,amssymb} 
\usepackage{color}
\newcommand{\etal}{\textit{et al}.}

\newcommand{\ie}{\textit{i}.\textit{e}.}
\newcommand{\eg}{\textit{e}.\textit{g}.}
\usepackage{sistyle}
\SIthousandsep{,}
\DeclareMathOperator*{\argmax}{arg\,max}

\usepackage{multirow}
\SIthousandsep{,}
\usepackage{sistyle}
\usepackage{booktabs}
\usepackage{graphics}
\usepackage{adjustbox}
\usepackage[colorlinks,linkcolor=blue]{hyperref}
\usepackage{subcaption}
\makeatletter
\renewcommand\normalsize{%
\@setfontsize\normalsize\@xpt\@xiipt
\abovedisplayskip 2\p@ \@plus2\p@ \@minus5\p@
\abovedisplayshortskip \z@ \@plus3\p@
\belowdisplayshortskip 6\p@ \@plus3\p@ \@minus3\p@
\belowdisplayskip \abovedisplayskip
\let\@listi\@listI}
\makeatother

\begin{document}
\pagestyle{headings}
\mainmatter
\def\ECCVSubNumber{6277}  

\title{Visual Relation Grounding in Videos} 

%
\author{Junbin Xiao\inst{1}\and
Xindi Shang\inst{1}\and
Xun Yang\inst{1}\and
Sheng Tang\inst{2}\and
Tat-Seng Chua\inst{1}}

\institute{Department of Computer Science, National University of Singapore, Singapore \email{\{junbin,shangxin,chuats\}@comp.nus.edu.sg, xunyang@nus.edu.sg} \and
Institute of Computing Technology, Chinese Academy of Sciences, China
\email{ts@ict.ac.cn}}
\maketitle

\begin{abstract}
In this paper, we explore a novel task named visual Relation Grounding in Videos (vRGV). The task aims at spatio-temporally localizing the given relations in the form of \emph{subject-predicate-object} in the videos, so as to provide supportive visual facts for other high-level video-language tasks (\eg, video-language grounding and video question answering). The challenges in this task include but not limited to: (1) both the subject and object are required to be spatio-temporally localized to ground a query relation; (2) the temporal dynamic nature of visual relations in videos is difficult to capture; and (3) the grounding should be achieved without any direct supervision in space and time. To ground the relations, we tackle the challenges by collaboratively optimizing two sequences of regions over a constructed hierarchical spatio-temporal region graph through relation attending and reconstruction, in which we further propose a message passing mechanism by spatial attention shifting between visual entities. Experimental results demonstrate that our model can not only outperform baseline approaches significantly, but also produces visually meaningful facts to support visual grounding. (Code is available at \href{https://github.com/doc-doc/vRGV}{https://github.com/doc-doc/vRGV}).
\end{abstract}

\section{Introduction}
Visual grounding aims to establish precise correspondence between textual query and visual contents by localizing in the images or videos the relevant visual facts depicted by the given language. It was originally tackled in language-based visual fragment-retrieval \cite{hu2016natural,karpathy2015deep,karpathy2014deep}, and has recently attracted widespread attention as a task onto itself. While lots of the existing efforts are made on referring expression grounding in static images \cite{hu2017modeling,liu2019adaptive,mao2016generation,nagaraja2016modeling,rohrbach2016grounding,yu2018mattnet,yu2016modeling,zhang2018grounding}, recent research attempts to study visual grounding in videos by finding the objects either in individual frames \cite{huang2018finding,shi2019not,zhou2018weakly} or in video clips spatio-temporally \cite{balajee2018object,chen2019weakly,zhang2020does}. Nonetheless, all these works focus on grounding in videos the objects depicted by natural language sentences. Although the models have shown success on the corresponding datasets, they lack transparency to tell which parts of the sentence help to disambiguate the object from the others, and thus hard to explain whether they truly understand the contents or just vaguely learn from data statistics. Furthermore, the models fail to effectively reason about the visual details (\eg, relational semantics), and thus would generalize poorly on unseen scenarios.

\begin{figure*}[ht]
\centering
  \includegraphics[width=0.9\textwidth]{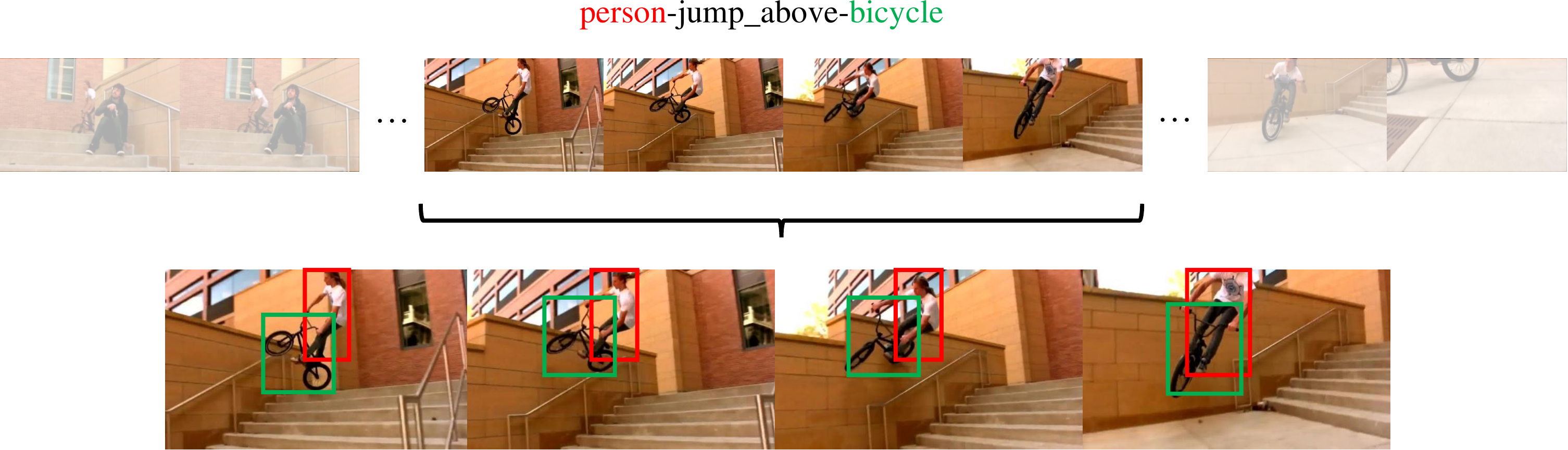}
  \caption{Illustration of the vRGV task. For the query relation \emph{person-jump\_above-bicycle} and an untrimmed video containing the relation, the objective is to find a video segment along with two trajectories corresponding to the subject (red box) and object (green box) that match the query relation.}
  \label{fig:introduction}
\end{figure*}

To achieve understandable visual analytics, many works have showed the importance of comprehending the interactions and relationships between objects \cite{hu2017modeling,krishna2017visual,shang2017video}. To this end, we explore explicit relations in videos by proposing a novel task of visual Relation Grounding in Videos (vRGV). The task takes a relation in the form of \emph{subject-predicate-object} as query, and requires the models to localize the related visual subject and object in a video by returning their trajectories. As shown in Fig.~\ref{fig:introduction}, given a query relation \emph{person-jump\_above-bicycle} and an untrimmed video containing that relation, the task is to find in the video the subject (person in white) and object (bicycle) trajectory pairs that hold the query relationship \emph{jump\_above}.\footnote{The word 'untrimmed' is regarding to relation. We refer to relation as the complete triplet \emph{subject-predicate-object}, and relationship as the \emph{predicate} only.} Considering that annotating fine-grained region-relation pairs in videos is complicated and labor-intensive \cite{shang2019annotating}, we define the task as weakly-supervised grounding where only video-relation correspondences are available. That is, the models only know that a relation exists but do not know where and when it is present in the video during training.

vRGV retains and invokes several challenges associated with video visual relation and visual grounding. First, unlike existing video grounding \cite{balajee2018object,chen2019weakly} which is to localize a specific object according to its natural language description, in visual relation grounding, the models are required to jointly localize a pair of visual entities (subject and object) conditioned on their local relationships. Second, unlike a coarsely global description, relations are more fine-grained and change over time, \ie, even a same object would have different relationships with different objects at different time. For example, 
the enduring relations (\eg, \emph{person-drive-car}) may exist for a long time but the transient ones (\eg, \emph{person-get\_off-car}) may disappear quickly. Besides, static relationships like spatial locations and states (\eg,  \emph{hold} and \emph{hug}) can be grounded at frame level, whereas the dynamic ones such as \emph{lift\_up} and \emph{put\_down} can only be grounded based on a short video clip. Such dynamic nature of relations in videos will cause great challenge for spatio-temporal modeling. 
Third, the requirement for weakly-supervision also challenges the models to learn to ground the relation without reliance on any spatial (bounding boxes) and temporal (time spans) supervisions.

To address the above challenges, we devise a model of video relation grounding by reconstruction. 
Concretely, we incorporate a query relation into a given video which is modeled as a hierarchical spatio-temporal region graph, for the purpose of identifying the related subject and object from region proposals in multi-temporal granularity.
For weakly-supervised video relation grounding during training, we optimize the localization by reconstructing the query relation with the subject and object identified by textual clues and attention shift. During inference, we dynamically link the subject and object which contribute most to the re-generated relations into trajectories as the grounding result. Our insight is that visual relations are data representations of clear structural information, and there is a strong cross-modal correspondence between the textual subject-object and the visual entities in the videos. Thus, it is reasonable to ground the relevant subject and object by re-generating the relation. 

Our main contributions are: (1) we define a novel task, visual Relation Grounding in Videos (vRGV), to provide a new benchmark for the research on video visual relation and underpin high-level video-language tasks; (2) we propose an approach for weakly-supervised video relation grounding, by collaboratively optimizing two sequences of regions over a hierarchical spatio-temporal region graph through relation attending and reconstruction; and (3) we propose a novel message passing mechanism based on spatial attention shifting between visual entities, to spatially pinpoint the related subject and object.



\section{Related Work}
In this section, we briefly recap the history in visual relation, visual grounding and video modeling, which are either similar in spirit to the task definition or technically relevant to our approach.

\textbf{Visual Relation}.
Early attempts on visual relation either leveraged object co-occurrence and spatial relationships for object segmentation \cite{galleguillos2008object}, or focused on human-centric relationships for understanding human-object interactions \cite{yao2010modeling}. Recently, many works started to study visual relations as a task onto itself to facilitate cognitive visual reasoning. Lu \etal~\cite{lu2016visual} firstly formulated visual relations as three separated parts of \emph{object\_1-predicate-object\_2}, and classified visual relationships as spatial, comparative, preposition and verb predicates. Krishna~\etal~\cite{krishna2017visual} formalized visual relations as a scene graph for image structural representation, in which visual entities are corresponding to nodes and connected by edges depicted by object relationships. Shang \etal~\cite{shang2017video} introduced visual relations from images to videos (video scene graph). Apart from the relations in static images, they added relationships that are featured with dynamic information (\eg, \emph{chase} and \emph{wave} ), so as to emphasize spatio-temporal reasoning of fine-grained video contents. According to their definition, a valid relation in videos requires both the subject and object to appear together in each frame of a certain video clip. 

While a handful of works have successfully exploited relations to improve visual grounding \cite{hu2017modeling} and visual question answering \cite{lu2018r}, relation as an independent problem is mostly tackled in the form of detection task and the advancements are mostly made in the image domain \cite{liang2018visual,lu2016visual,Zhang2019large}. In contrast, relation as a detection task in video domain has earned little attention, partly due to the great challenges in joint video object detection and relation prediction with insufficient video data \cite{qian2019video,tsai2019video}. In this paper, instead of blindly detecting all visual objects and relations in videos, we focuses on the inverse side of the problem by spatio-temporally grounding a given relation in a video.

\textbf{Visual Grounding}.
Visual grounding has emerged as a subject under intense study in referring expression comprehension \cite{hu2017modeling,krishna2018referring,mao2016generation,nagaraja2016modeling,rohrbach2016grounding,yu2018mattnet,yu2016modeling,zhang2018grounding}. Mao \etal~\cite{mao2016generation} first explored referring expression grounding by using the framework of Convolutional Neural Network (CNN) and Long-Short Term Memory (LSTM) network~\cite{hochreiter1997long} for image and sentence modeling. They achieved grounding by extracting region proposals and then finding the region that can generate the sentence with maximum posterior probability. Similarly, Rohrbach \etal~\cite{rohrbach2016grounding} explored image grounding by reconstruction to enable grounding in a weakly-supervised scenario. 
Krishna \etal~\cite{krishna2018referring} explored referring relationships in images by iterative message passing between subjects and objects. While these works focus on image grounding, more recent efforts \cite{balajee2018object,chen2019weakly,huang2018finding,shi2019not,yamaguchi2017spatio,zhang2020does,zhou2018weakly} also attempted to ground objects in videos. Zhou \etal~\cite{zhou2018weakly} explored weakly-supervised grounding of descriptive nouns in separate frames in a frame-weighted retrieval fashion. Huang \etal~\cite{huang2018finding} proposed to grounding referring expression in temporally aligned instructional videos. Chen \etal~\cite{chen2019weakly} proposed to perform spatio-temporal object grounding with video-level supervision, which aims to localize an object tube described by a natural language sentence. They pre-extracted the action tubes, and then rank and return the tube of maximal similarity with the query sentence. Instead of grounding a certain object in trimmed videos by a global object description \cite{chen2019weakly}, we are interested in localizing a couple of objects conditioned upon their relationships in untrimmed videos, which is more challenging and meaningful in reasoning real-world visual contents.


\textbf{Video Modeling}.
Over the decades, modeling the spatio-temporal nature of video has been the core of research in video understanding. Established hand-crafted feature like iDT \cite{wang2013action} and deep CNN based features like C3D \cite{tran2015learning}, two-Stream \cite{simonyan2014two} and I3D \cite{carreira2017quo}, have shown their respective strengths in different models. However, all these features mainly capture motion information in a short time interval (\eg, 16 frames as the popular setting in C3D). To enable both long and short-term dependency capturing, researchers \cite{venugopalan2015sequence,yue2015beyond} also attempted to model the video as an ordered frame sequence using Recurrent Neural Networks (RNNs). While RNN can deal with dynamic video length in principle, it was reported that the preferable number of frames with regard to a video should be ranged from 30 to 80 \cite{pan2016hierarchical,venugopalan2015sequence}. As a result, Pan \etal~\cite{pan2016hierarchical} further proposed a hierarchical recurrent neural encoder to achieve temporal modeling in multiple granularity. Yet, they focused on generating a global description of the video by extracting frame-level CNN feature, which can hardly be applied to relation understanding where fine-grained regional information is indispensable. Recently, there is a tendency of modeling videos as spatio-temporal graphs \cite{jain2016structural,qian2019video,tsai2019video,wang2018videos}, where the nodes correspond to regions and edges to spatial-temporal relationships. Nonetheless, all of them model the video as a flat and densely connected graph. Instead, we retain the temporal structure (ordered frames and clips) of videos by modeling it as a hierarchical spatio-temporal region graph with sparse directed connections. 

\section{Method}
Recall that our goal is to ground relations in the given videos, which is formulated by giving a relation coupled with videos containing that relation, and returning two trajectories for each video, corresponding to the subject and object participating in the query relation\footnote{If there are more than one instances that match the query relation in a video, it is a correct grounding by returning any one of them.}. We formally define the task as follows.

\textbf{Task Definition:} Given a set of query relations in form of $\mathcal{R}=\{<S-P-O>\}$ and a set of untrimmed videos $\mathcal{V}$ (where $S$, $P$, $O$ denote the $subject$, $predicate$ and $object$ respectively), and each specific query relation $\mathcal{R}_i$ is coupled with several videos from $\mathcal{V}$ which contain that relation, the task is to spatio-temporally localize in the videos the respective subjects and objects by returning their trajectories $T_s, T_o$. The trajectory $T$ is given by a sequence of bounding boxes tied to a certain visual entity across a video segment. For weakly-supervised grounding, there is no spatial (bounding box) and temporal supervisions (time spans) from the dataset during training. 

\subsection{Solution Overview}
Given a video of $N$ frames, we first extract $M$ region proposals for each frame. Thus, a video can be represented by a set of regions $V=\{B_{i,j} ~\vert~ i\in [1,N],~ j\in [1,M]\}$, and a trajectory $T=\{B_i~|~i\in [k,l],~k\in[1, N],~l\in[k, N]\}$ can be a sequence of bounding boxes in the video. Our approach will learn to ground a given relation $R$ by finding two trajectories $T_s, T_o$ that indicate the subject and object of the relation. According to the task definition, we resolve the problem by maximizing the following posterior probability:
\begin{equation}
\label{equ:framework}
    T_s^*, T_o^* = \argmax_{T_s,~T_o} \mathit{P}(R~|~T_s,~T_o)*\mathit{P}(T_s,~ T_o~|~V,~R),
\end{equation}
where $\mathit{P}(T_s,~ T_o~|~V,~R)$ aims to attend to the most relevant trajectories in $V$ given the relation $R$, and $\mathit{P}(R~|~T_s,~T_o)$ attempts to reconstruct the same relation $R$ based on the relevant trajectories it attended to. During inference, our approach will output the trajectories ($T_s^*$ and $T_o^*$) that contribute mostly to the re-generated relation to accomplish grounding. 
\begin{figure*}[t]
\begin{center}
\includegraphics[width=0.9\linewidth]{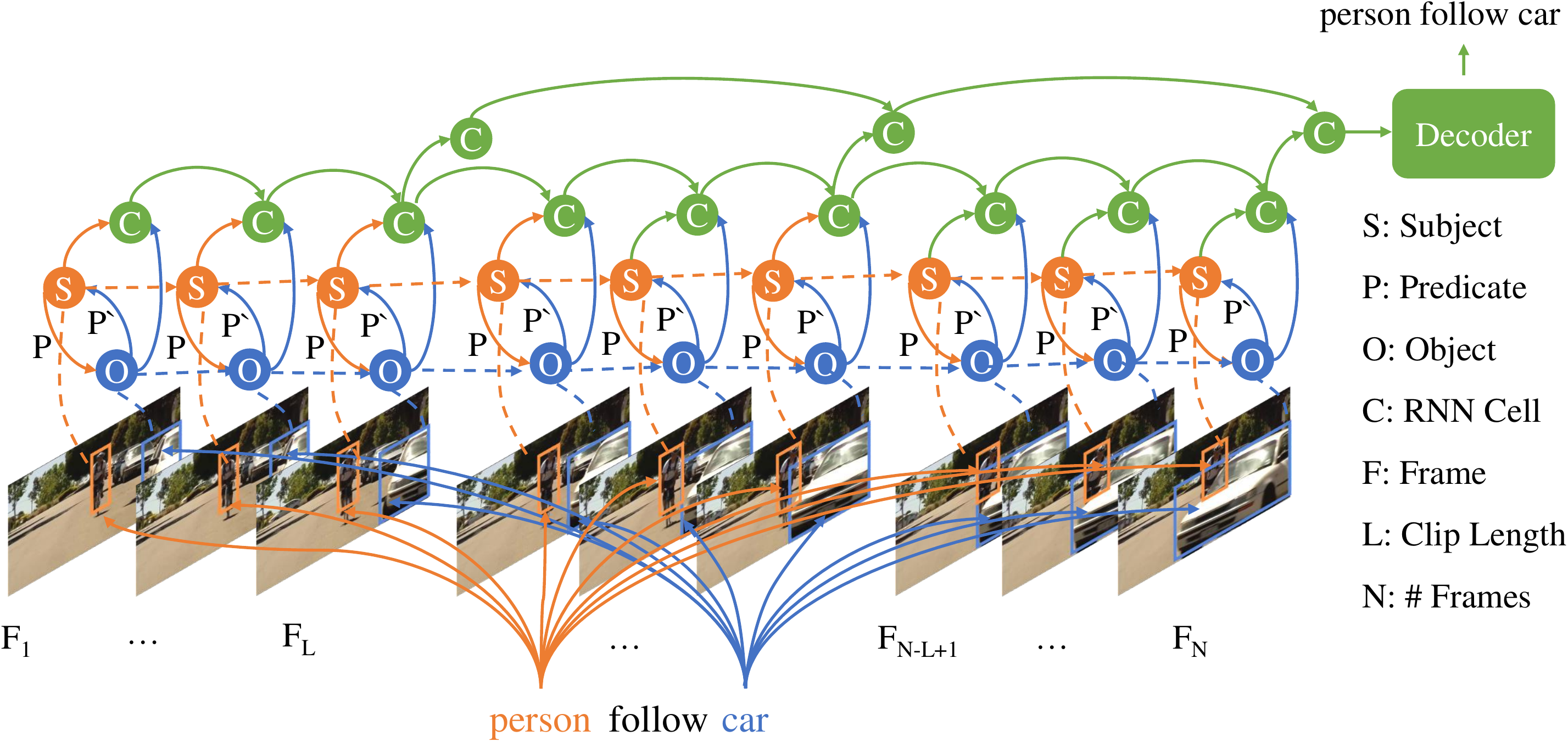}
\end{center}
\caption{Schematic diagram of video relation grounding by reconstruction. The model takes the query relation as guidance to pinpoint regions of subjects and objects over the hierarchical spatio-temporal region graph, where the regions correspond to nodes which are spatially connected by visual relationships and are temporally connected by hierarchical RNN over different frames and clips.}
\label{fig:method}
\end{figure*}

The key idea of our approach is to ground the relation through reconstruction by capturing the intuition that there is a clear correspondence between subject-object in textual relation and visual instances. However, unlike image grounding \cite{rohrbach2016grounding} which can directly model and return the region proposals, effective trajectory proposals are unavailable in this task due to the complicated dynamics of the relations in videos. To achieve the optimization in Equ.~(\ref{equ:framework}), we model the trajectory proposals implicitly over a hierarchical spatio-temporal region graph and online-optimize them through relation attending and reconstruction.
As shown in Fig.~\ref{fig:method}, two sequences of regions corresponding to the query subject and object will be identified from the video (region graph) and jointly embedded into the final graph representation which will be fed to the decoder to reconstruct the relation. The reconstruction loss from the decoder will be back-propagated to the graph encoder, to penalize the incorrect object pairs with respect to the relation. During inference, we dynamically link the regions which response significantly to the reconstructed relations into explicit trajectories to accomplish the grounding, where the importance of the regions are determined by spatial and temporal attention over the hierarchical space-time region graph. In this way, our model can in principle ground visual relations in multi-temporal granularity without the bottleneck of off-line trajectory proposal extraction \cite{chen2019weakly}. 

We next elaborate how to learn to spatio-temporally attend to the correct sequences of regions for a given relation, and then how to obtain the final trajectories based on the attention values, to accomplish the grounding.
 
\subsection{Message Passing by Attention Shifting}
\textbf{Spatial Attention}. This unit takes as input all the region proposals $B=\{B_j~\vert~j\in[1,M] \}$ in a frame and the query relation $R=<S-P-O>$. It learns two spatial attentions ($\alpha_s^{M\times 1},~\alpha_o^{M\times 1}$) corresponding to the subject and object. Concretely, the spatial attention unit (SAU) is formulated as:
\begin{equation}
\begin{split}
    \alpha_s = SAU(~f(B), ~g(S)~),~
    \alpha_o = SAU(~f(B), ~g(O)~),
\end{split}
\end{equation}
in which $g(\cdot)$ returns the textual word feature for subject (S) or object (O), and is achieved by embedding the respective GloVe \cite{pennington2014glove} vector: $g(S) = Emb(GloVe(S))$. Besides, $f(\cdot)$ means feature extraction for the region proposals. In our implementation, we utilize several kinds of feature related to the object appearances, relative locations and sizes, which are not only important in visual relation understanding, but also crucial in identifying the same object in different frames.

The object appearance is captured by the ROI-aligned feature from object detection models, \ie, $f_{app}=CNN(B_j)$. The object relative location and size are useful to identify the spatial and comparative relationships, and are given by $f_B=[\frac{x_{min}}{W},\frac{y_{min}}{H}, \frac{x_{max}}{W}, \frac{y_{max}}{H}, \frac{area}{W*H}]$, in which $W, H$ are respectively the width, height of the frame and $area$ is the area of the bounding box represented by the top-left $(x_{min}, y_{min})$ and bottom-right $(x_{max}, y_{max})$ coordinates. The final feature of a region proposal $f(B_j)$ is thus obtained by element-wise addition of the transformed visual appearance feature $f_{app}$ and bounding box feature $f_B$. The transform operation is achieved by linear mapping with ReLU activation.

We take the subject $S$ as an example to introduce how to obtain the representation $f_s$ and attention distribution $\alpha_s$ for it (similar way to obtain $\alpha_o$ and $f_o$ for object $O$). Given the textual subject representation $g(S)$ and each region proposal $f(B_j)$, the attention score $s_j$ is obtained by
\begin{equation}
\label{equ:spatt1}
    s_j = W_2tanh(W_1[f(B_j),~g(S)]+b_1),
\end{equation}
in which $W_1,b_1,W_2$ are model parameters. Then, the attention distribution over different region $B_j$, and the final representation for subject $S$ are give by
\begin{equation}
\label{equ:spatt2}
    \alpha_{s_j} = softmax(s_j) = \frac{exp(s_j)}{\sum_{z=1}^M exp(s_z)},~
    f_s = \sum_{j=0}^M \alpha_{s_j}f(B_j).
\end{equation}

\noindent\textbf{Attention Shifting}. 
Although the aforementioned attention unit is capable of identifying the subjects and objects semantically related to the query relations, they are not necessarily the exact visual entities that hold the relationships. Take the instance in Fig.~\ref{fig:method} as an example, there is one \emph{person} but several \emph{cars} on the street, and only one \emph{car} match the relationship \emph{follow} with the \emph{person}. It is the relationship \emph{follow} that helps to disambiguate the \emph{car} of interest from the other \emph{cars}. Another intuition is that, given the subject \emph{person} and relationship \emph{follow}, the searching space of the object \emph{car} can be narrowed to the areas in front of the \emph{person}, and vice verse. 

As shown in Fig~\ref{fig:method}, we capture these insights by modeling the relationships as attention shifting (message passing) between the visual entities, so as to accurately pinpoint the subject and object participating in the query relation. Specifically, we learn two independent transfer matrices $W_{so}$ and $W_{os}$ tied to the forward relationship ($P$, message from subject to object) and backward relationship ($P'$, message from object to subject) respectively. 
\begin{equation}
\begin{split}
   f_{so}  = ReLU(W_{so}\alpha_s),~
   f_{os}  = ReLU(W_{os}\alpha_o).
\end{split}
\end{equation}
The transferred location feature from subject (object) to object (subject) will be added to the attention based object (subject) feature:
\begin{equation}
\begin{split}
   f_s  = f_s + f_{os},~
   f_o  = f_o + f_{so}.
\end{split}
\end{equation}
Finally, the subject and object representations will be concatenated and transformed to obtain the node input at time step i, \ie, $f_i = W_3([f_s, f_o]) + b_3$, where $W_3$, $b_3$ are learnable parameters.

\subsection{Hierarchical Temporal Attention}
To cope with the temporal dynamics of video relations, we devise two relation-aware hierarchical temporal attention units $TAU_1$ and $TAU_2$ (which work in a way similar to $SAU$) over the frames and clips respectively. As shown in Fig. \ref{fig:method}, a video is firstly divided into $H=\frac{N}{L}$ short clips of length $L$. The frame-wise temporal attention $\beta^{l1}$ (of dimension $N$) is obtained by
\begin{equation}
    \beta^{l1} = TAU_1(f^{l1}, ~f^{l2}_H),
\end{equation}
in which $f^{l1}$ denotes the sequence of frame-wise feature, and is achieved by sequence modeling the subject and object concatenated feature $f$
\begin{equation}
    f^{l1}_i = LSTM_{l1}(f_{1,\cdots,i}),~i\in[1,N].
\end{equation}
Besides, $f^{l2}_H$ denotes the output at the last time step of the clip-level neural encoder, which is obtained by 
\begin{equation}
    f^{l2}_H = LSTM_{l2}((\beta^{l2}_i f^c_i)_{i=1,\cdots, H}),
\end{equation}
where $f^c$ denotes the sequence of clip-level inputs which are obtained by selecting the output of the first layer of LSTM at every L steps, \ie, $f^c=\{f^{l1}_i~\vert~ i\in\{1, L, \cdots, N\}\}$. $\beta^{l2}$ (of dimension $H$) is the clip-level temporal attention distribution, and is obtained by
\begin{equation}
    \beta^{l2} = TAU_2(f^c, ~f_R),\\
\end{equation}
where $f_R$ denotes the query relation which is obtained by concatenating the GloVe feature of each part in the relation (average for phrase) and further transformed to the same dimension space as feature vectors in $f^c$.

\subsection{Train and Inference}
\label{sec:train-inference}
During training, we drive the final graph embedding by an attention-guided pooling of the node representations across the video, \ie, $feat_v=\sum \beta^{l1}f^{l1}$. The graph embedding will be fed to the decoder to re-generate the query relation. The decoder part of our model is similar to \cite{rohrbach2016grounding} by treating the relation as a textual phrase and reconstruct it by a single LSTM layer. The model was trained with the cross-entropy loss
\begin{equation}
L_{rec} = -\frac{1}{n_{vr}}\sum_{n=1}^{n_{vr}}\sum_{t=1}^{n_w} log P(R_t|R_{0:t-1}, ~feat_v),
\end{equation}
where $R_t$ denotes the $t^{th}$ word in the relation. $n_{vr}$ and $n_w$ denote the number of video-relation samples and number of words in the relation respectively.

During inference, we base on the learned spatio-temporal attention values to achieve the relation-aware trajectories.
First, we temporally threshold to obtain a set of candidate sub-segments for each relation-video instance in the test set. The segments are obtained by grouping the successive frames in the remaining frame set after thresholding with value $\sigma$, \ie,
$
B=\{B_{i,1:M}|\beta_i>=\sigma\}, 
$
in which $B_i$ denotes regions in frame $i$. $\beta$ is temporal attention value obtained by $\beta=\beta^{l1}+\beta^{l2}$. Note that the clip-level attention value will be propagated to all frames belonging to that clip. (Refer to appendix for more details.)

Then, for each sub-segment, we define a linking score $s(B_{i,p}, B_{i+1,q})$ between regions of successive frames (after sampling) 
\begin{equation}
\label{equ:viterbi}
    s(B_{i,p}, B_{i+1,q}) = \alpha_{i,p} + \alpha_{i+1,q} + \lambda \cdot IoU(B_{i,p}, B_{i+1,q}),
\end{equation}
where $\alpha$ is the spatial attention value, it can be $\alpha_s$ (subject) or $\alpha_o$ (object) depending on the linking visual instances. $IoU$ denotes the overlap between two bounding boxes, and $\lambda=\frac{1}{D}$ is a balancing term related to the distance ($D\in[1,10]$) of the two successive frames. By defining $\lambda$, we trust more on the attention score when the distance between the two frames are larger. Our idea is to link the regions which response strongly to the subject or object, and their spatial extent overlaps significantly. The final trajectory can thus be achieved by finding the optimal path over the segment  
\begin{equation}
T^*=\argmax_{T}\frac{1}{K-1}\sum_{i=1}^{K-1} s(B_{i,p}, B_{i+1,q}),
\end{equation}
where $T$ is a certain linked region sequence of length $K$ for the subject or object. Similar to \cite{gkioxari2015finding}, we solve the optimization problem using Viterbi algorithm. Finally, the linking scores associated with the subject and object are averaged to obtain the score for the corresponding sub-segment. The grounding is achieved by returning the segment (subject-object trajectory pair) of maximal score.

\section{Experiments}
\label{sec:exp}
\subsection{Dataset and Evaluation}
We conduct experiments on the challenging video relation dataset ImageNet-VidVRD \cite{shang2017video}. It contains 1000 videos selected from ILSVRC-VID \cite{ILSVRC15}, and is annotated with over \num{30000} relation instances covering 35 object classes and 132 predefined predicates. Our preliminary investigation shows that over 99\% of relations do not appear throughout the video, with 92\% (67\%) appearing in less than 1/2 (1/5) length of the video, and the shortest relation only exists in 1 second, while the longest relation lasts for 40 seconds. Besides, each video contains 2 to 22 objects (3 on average), excluding those un-related objects which also matter due to the weakly-supervised setting. The dataset statistics are listed in Table \ref{tab:dataset}, others details are given in the appendix. Note that the object trajectories are provided but are not used during training.  
\begin{table*}[t]
\begin{center}
\caption{Statistics of ImageNet-VidVRD. }
\label{tab:dataset}
\begin{tabular}{llccccc}
\toprule
\multicolumn{2}{c}{Dataset} & \#Videos & \#Objects & \#Predicates & \#Relations & \#Instances \\
\midrule
\midrule
\multirow{2}{*}{\shortstack{ImageNet- \\ VidVRD \cite{shang2017video}}}  & Train  & 800 & 35 & 132 &2961 & \num{25917} \\
                  & Val &  200 & 35 & 132 & 1011 & \num{4835} \\
\bottomrule
\end{tabular}
\end{center}
\end{table*}

We report accuracy (in percentage) as the grounding performance. Specifically, for each query relation, there might be one or more videos, with each having one or more visual instances corresponding to that relation. For a video-relation pair, a true-positive grounding is confirmed if the returned subject-object trajectory has an overlap ratio of larger than 0.5 with one of the ground-truth visual relation instances. The overlap is measured by the temporal Intersection over Union (tIoU), which is based on the average number of three different spatial IoU thresholds (\ie, sIoU = 0.3, 0.5 and 0.7). Aside from the joint accuracy for the whole relation ($Acc_R$), we also separately report the accuracy for the subject ($Acc_S$) and object ($Acc_O$) for better analysis of algorithms.
\subsection{Implementation Details}
For each video-relation instance, we uniformly sample N=120 frames from the video and further divide them into H=10 clips of length L=12 frames, and extract M=40 region proposals for each frame. We apply Faster R-CNN \cite{ren2015faster} with ResNet-101 \cite{he2016deep} as backbone (pretrained on MS-COCO~\cite{lin2014microsoft}) to extract the region proposals, along with the 2048-D regional appearance feature. The final dimension of each region representation is transformed to 256. For each word in the textual relation, we obtain the 300-D GloVe feature and then embed it to 256 dimension space. The hidden size for the encoder and decoder LSTM is 512. Besides, the models are trained using Adam \cite{kingma2014adam} optimization algorithm based on an initial learning rate of 1e-4 and batch size of 32. We train the model with maximal 20 epochs, and use dropout rate 0.2 and early stopping to alleviate over-fitting. During inference, we first obtain the spatial and temporal grounding results on the basis of all the sampled frames, and then propagate the adjacent bounding boxes to the missing frames based on linear interpolation (see appendix for details). The temporal attention threshold is set to 0.04 , and is greedily searched on a validation split of the training data.
\subsection{Compared Baselines}
As there is no existing method for the vRGV task, we adapt several related works as our baselines. (1) \textbf{Object Co-occurrence}, it was applied in \cite{krishna2018referring} for referring relationship in images. We can equivalently achieve it in the video scenario by removing all the predicates in our model and only grounding the two categories. This baseline is to study how much the object co-occurrence will contribute to the relation grounding performance. (2) \textbf{Trajectory Ranking}. We adapt the method proposed in video language grounding (namely WSSTG \cite{chen2019weakly}) to the relation scenario. This can be achieved by regarding the relation as a natural language sentence and transforming grounding to sentence matching with the pre-extracted object tubes.
Specifically, we consider two implementation variants: (a) ${\bf V_1}$, which optimizes the similarity between each trajectory proposal and the query relation during training, and outputs the top-2 ranked trajectories as the grounded subject and object during inference; and (b) ${\bf V_2}$, which concatenates the trajectories pair-wisely to compare their similarity with the query sentence during training, and returns the top-1 trajectory-pair as grounded results during inference. (More details can be found in the appendix.)
\subsection{Result Analysis}
Table \ref{tab:results} shows the performance comparisons between our approach and the baselines, where vRGV* (similar for Co-occur*) denotes our model variant that greedily links the regions of maximal attention score in each frame by setting $\lambda$ in Equ.~(\ref{equ:viterbi}) to 0. We conduct this experiment to validate that our model is capable of learning the object identity across different frames, because we implicitly model and optimize the trajectories on the spatio-temporal graph. When the model is complemented with explicit object locations during post-linking, it can achieve better performances as shown in the bottom row.
\begin{table*}[t]
\begin{center}
\caption{Results of visual relation grounding in videos. We add bold and underline to highlight the best and second-best results under each metric respectively.}
\label{tab:results}
\scalebox{0.89}{
\begin{tabular}{l|ccc|ccc|ccc|ccc}
\toprule
\multirow{2}{*}{Methods} & \multicolumn{3}{c|}{sIoU=0.3} & \multicolumn{3}{c|}{sIoU=0.5} & \multicolumn{3}{c|}{sIoU=0.7} & \multicolumn{3}{c}{Average}\\
& $Acc_S$ & $Acc_O$ & $Acc_R$ & $Acc_S$ & $Acc_O$ & $Acc_R$ & $Acc_S$ & $Acc_O$ & $Acc_R$ & $Acc_S$ & $Acc_O$ & $Acc_R$\\
\midrule
\midrule
T-Rank $V_1$ \cite{chen2019weakly} & 33.55 & 27.52 & 17.25 & 22.61 & 12.79 & 4.49 & 6.31 & 3.30 & 0.76 & 20.27 & 10.68 & 3.99\\
T-Rank $V_2$ \cite{chen2019weakly} & 34.35 & 21.71 & 15.06 & 23.00 & 9.18 & 3.82 & 7.06 & 2.09 & 0.50 & 20.83 & 7.35 & 3.16\\
Co-occur* \cite{krishna2018referring} & 27.84 & 25.62 & 18.44 & 23.50 & 20.40 & 13.81 & 17.02 & 14.93 & 7.29 & 22.99 &19.33  & 12.80\\
Co-occur \cite{krishna2018referring} & 31.31 & 30.65 & 21.79 & 28.02 & 27.69 & 18.86 & \underline{21.99} & \underline{21.64} & \underline{13.16} & 25.90 & 25.23 & 16.48\\
vRGV* (ours) & \underline{37.61} & \underline{37.75} & \underline{27.54} & \underline{32.17} & \underline{32.32} & \underline{21.43} & 21.34 & 21.02 & 10.62 & \underline{31.64} & \underline{30.92} & \underline{20.54}\\
vRGV~ (ours) & \textbf{42.31} & \textbf{41.31} & \textbf{29.95} & \textbf{37.11} & \textbf{37.52} & \textbf{24.77} & \textbf{29.71} & \textbf{29.72} & \textbf{17.09} & \textbf{36.77} & \textbf{36.30} & \textbf{24.58}\\
\bottomrule
\end{tabular}
}
\end{center}
\end{table*}

\begin{figure*}[t]
\centering
  \includegraphics[width=0.83\textwidth]{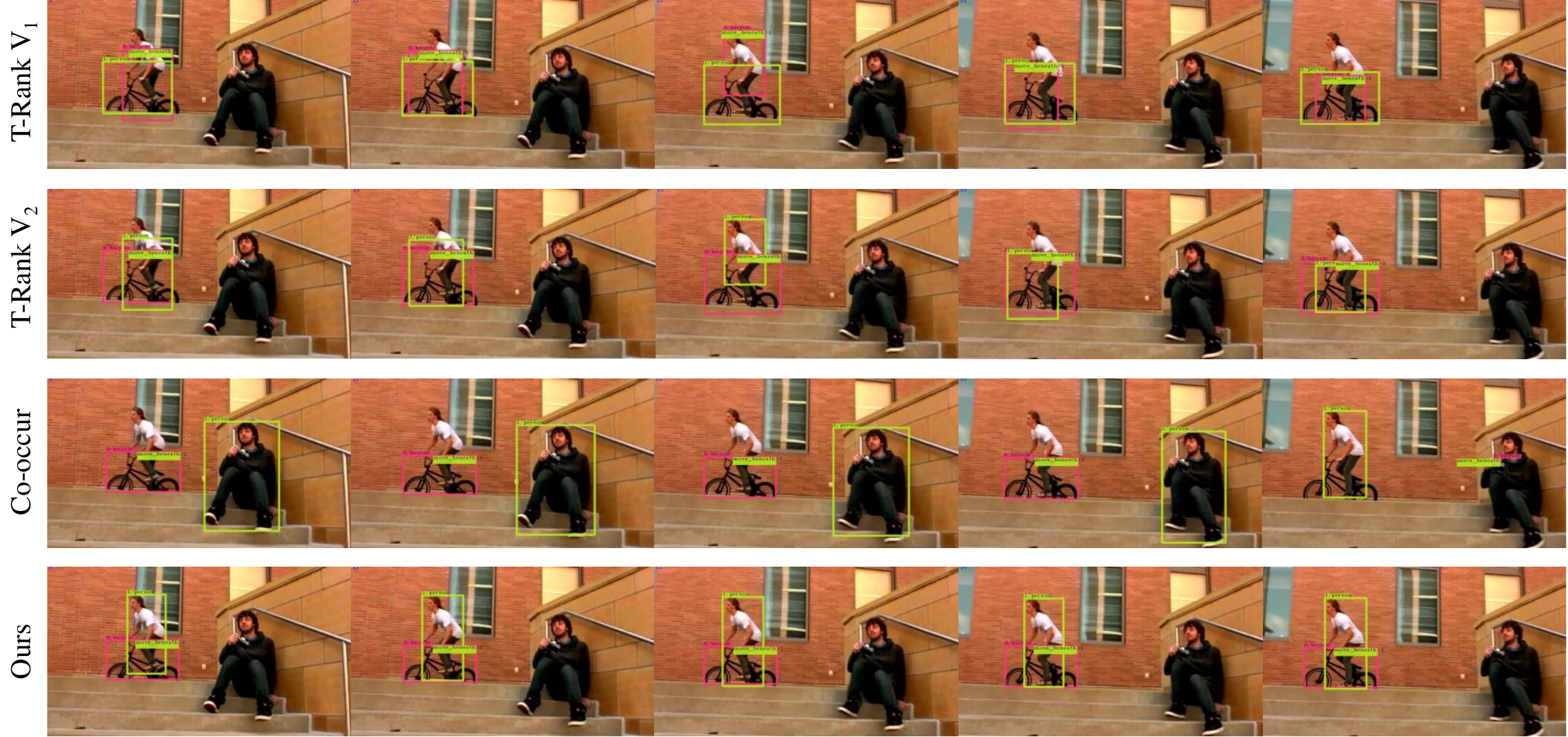}
  \caption{Qualitative results on the query relation \emph{\textcolor{red}{bicycle}-move\_beneath-\textcolor{green}{person}}.}
  \label{fig:comp}
\end{figure*}

From the results, we can see that our methods significantly outperform the baselines, and both methods adapted from WSSTG \cite{chen2019weakly} (\ie, T-Rank $V_1$ and T-Rank $V_2$) perform poorly on this task. We speculate the reasons are two folds: (1) the method in WSSTG is designed for single object grounding, they fail to jointly ground two visual entities and further to disambiguate between subject and object (see Fig.~\ref{fig:comp}). In our approach, we collaboratively optimize two sequences of objects on the spatio-temporal graph with relation attending and message passing mechanisms, and thus cope well with the joint grounding problem. This is supported by the observation that the two baselines \cite{chen2019weakly} obtain relatively closer results to ours on separate grounding accuracy ($Acc_S$), but much lower results than ours regarding the joint accuracy $Acc_R$; (2) The method in WSSTG aims for object grounding in trimmed video clips, and it pre-extracts relation agnostic object tube proposals and keeps them unchanged during training. In contrast, our approach enables online optimization of object trajectories regarding relation and post-generates relation-aware trajectory pairs. Thus, we can generate better trajectories tailored for relations. This is supported by the observation that the two baselines can get closer results to ours at a relatively lower overlap threshold, but their results degenerate significantly at higher thresholds. (Please also refer to our results on different temporal overlap thresholds shown in the appendix.)

Another observation is that T-Rank $V_2$ performs better than T-Rank $V_1$ in grounding the subject ($Acc_S$), but gets much worse results in terms of object ($Acc_O$) and hence acts poorly on the joint grounding results for relations ($Acc_R$). This indicates that the two objects in the top-ranked trajectory pair usually do not correspond to the subject and object mentioned in the sentence, and they are more likely the redundant proposals of the main objects. As shown in Fig.~\ref{fig:comp}, the model T-Rank $V_2$ can successfully find the subject \emph{bicycle}, but fails to localize the object \emph{person}. We think the reason is that WSSTG \cite{chen2019weakly} is oriented for grounding the main object in a natural language sentence (\ie, the subject), and when concatenating the representations of two trajectories, it can enhance the representation for the main object, but not sure for other supportive objects. According to the results, it even confuses the model and thus jeopardizes the grounding performance for the object ($Acc_O$). 

Relatively, the co-occurrence baseline performs better than \cite{chen2019weakly} on this task. Yet, its performances are still worse than ours (Co-occur* v.s. vRGV* and Co-occur v.s. vRGV). This demonstrates that the ``predicate'' in the relation is crucial in precisely disambiguating the subjects and objects. As shown in Fig.~\ref{fig:comp}, the occurrence baseline wrongly grounds the person sitting as the object, whereas our method successfully grounds the object to the person riding the bicycle. We also note that the co-occurrence baseline beats our weak model variant under the metric with threshold 0.7 (Co-occur v.s. vRGV*), which in fact, shows the superiority of our overall framework for joint grounding of two objects. Also, it indicates the importance of object locations in generating better trajectories. 

\subsection{Model Ablations}
\begin{table*}[t]
\begin{center}
\caption{Model ablation results on ImageNet-VidVRD.}
\label{tab:aba_results}
\scalebox{0.93}{
\begin{tabular}{c|ccc|ccc|ccc|ccc}
\toprule
\multirow{2}{*}{Models} & \multicolumn{3}{c|}{sIoU=0.3} & \multicolumn{3}{c|}{sIoU=0.5} & \multicolumn{3}{c|}{sIoU=0.7} & \multicolumn{3}{c}{Average}\\
& $Acc_S$ & $Acc_O$ & $Acc_R$ & $Acc_S$ & $Acc_O$ & $Acc_R$ & $Acc_S$ & $Acc_O$ & $Acc_R$ & $Acc_S$ & $Acc_O$ & $Acc_R$\\
\midrule
\midrule
vRGV & \textbf{42.31} & \textbf{41.31} & \textbf{29.95} & \textbf{37.11} & \textbf{37.52} & \textbf{24.77} & \textbf{29.71} & \textbf{29.72} & \textbf{17.09} & \textbf{36.77} & \textbf{36.30} & \textbf{24.58}\\
w/o Msg & 34.72 & 33.23 & 23.96 & 31.60 & 29.15 & 19.43 & 22.56 & 21.36 & 11.78 & 29.41 & 27.46 & 17.63\\
w/o Clip & \underline{41.08} & \underline{39.64} & \underline{27.15} & \underline{36.31} & \underline{35.05} & \underline{21.77} & \underline{28.19} & \underline{27.11} & \underline{13.72} & \underline{35.05} & \underline{34.03} & \underline{20.58}\\
w/o TAU & 32.99 & 32.76 & 20.34 & 22.36 & 19.99 & 7.61 & 15.29 & 13.27 & 4.83 & 21.75 & 19.26 & 7.06\\
\bottomrule
\end{tabular}
}
\end{center}
\end{table*}

\begin{figure*}[t]
\centering
  \includegraphics[width=0.9\textwidth]{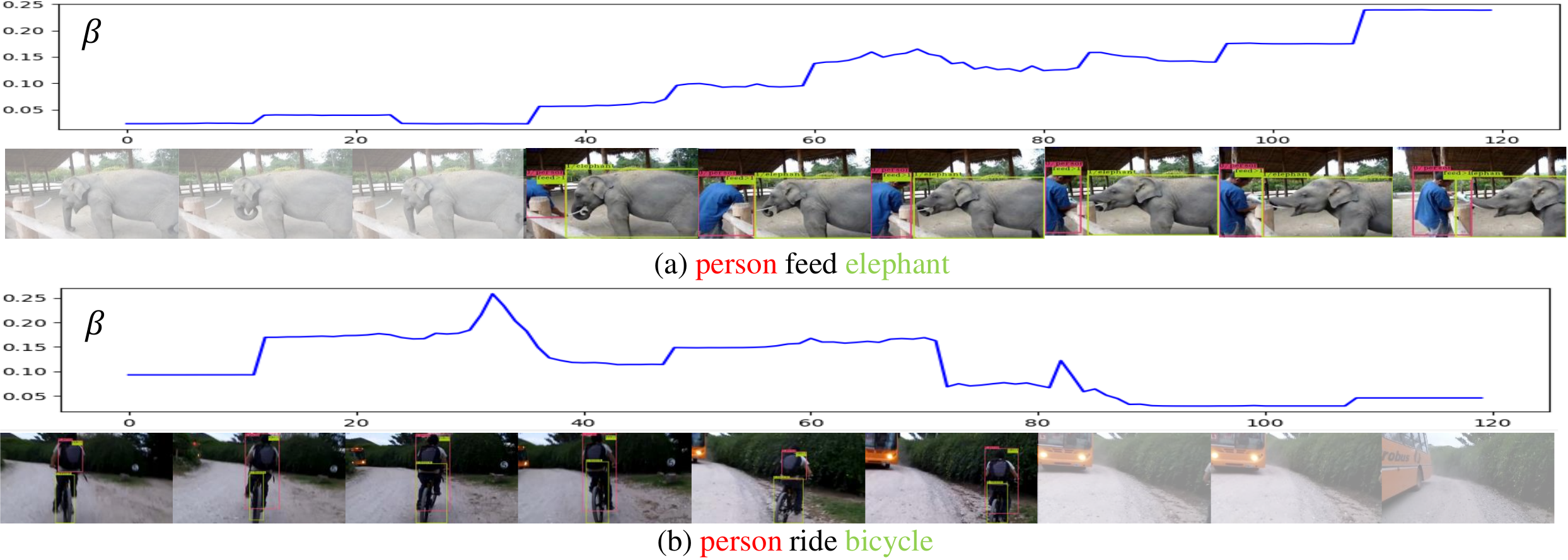}
  \caption{Qualitative results based on temporal threshold 0.04.}
  \label{fig:stu}
\end{figure*}
We ablate our model in Table \ref{tab:aba_results} to study the contribution of each component. 
Results in the 2nd row are obtained by removing the message passing module. We can see that the performance $Acc_R$ drops from 24.58\% to 17.63\%. This is mainly because the model without explicit message communication between subject and object cannot cope well with the scenario where there are multiple visual entities of same category present in the video. Another reason could be that the ablated model is weak in detecting objects under complex conditions (\eg, occlusion and blur) without the contextual information from their partners.

Results in the 3rd row are obtained by removing the clip-level attention. We do this ablation to prove the importance of the hierarchical structure of our model. Note that the temporal threshold $\sigma$ for this experiment is set to 0.0001, because we only have the frame-level attention $\beta^{l1}$. Comparing with results in the first row, we can see the results drop under all criteria without the hierarchical structure. When we further remove the frame-level attention (shown in the 4th row), and thus delete the whole temporal attention unit (TAU), the results degrade significantly from 24.58\% to 7.06\%. This is because our model will blindly link the objects throughout the video regardless of relation without the temporal grounding module. These findings demonstrate the importance of relation-aware temporal grounding in untrimmed videos.

We also analyze the temporal threshold $\sigma$ by changing it from 0.01 to 0.05, the corresponding results $Acc_R$ are 24.51\%, 24.76\%, 24.43\%, 24.58\% and 22.94\%. From the results, we can see that there is no significant difference when the threshold changes from 0.01 to 0.05, and the best result is achieved at the threshold of 0.02. We show some qualitative results in Fig.~\ref{fig:stu} based on temporal threshold of 0.04 (More results can be found in the appendix), from which we can see that our model can ground the subjects and objects in the videos when the query relations exist.

\subsection{Zero-shot Evaluation}
In this section, we analyze the models' capability of grounding the new (unseen during training) relation triplets. Specifically, in zero-shot relation grounding, we consider the case that the complete relation triplet is never seen, but their separate components (\eg, \emph{subject}, \emph{predicate} or \emph{object}) are known during training. For example, the model may have seen the relation triplets \emph{person-ride-bicycle} and \emph{person-run\_behind-car} during training, but it never knows \emph{bicycle-run\_behind-car}. As a result, we can find 432 relation instances of 258 unseen relation triplets in 73 videos from the test set.
\begin{table}[t]
\begin{center}
\caption{Results of zero-shot visual relation grounding.}
\label{tab:zeroresults}
\begin{tabular}{l|cccc}
\toprule
Methods & $Acc_S$ & $Acc_O$ & $Acc_R$ \\
\midrule\midrule
T-Rank $V_1$ \cite{chen2019weakly} & 4.05 & 4.08 & 1.37\\
T-Rank $V_2$ \cite{chen2019weakly} & 7.09 & 4.13 & 1.37\\
Co-occur \cite{krishna2018referring} & \underline{11.60} & \underline{10.99} & \underline{7.38}\\
vRGV (ours) & \textbf{18.94} & \textbf{17.23} & \textbf{10.27}\\
\bottomrule
\end{tabular}
\end{center}
\end{table}

As shown in Table \ref{tab:zeroresults}, our approach still outperforms the baselines in the zero-shot setting. We attribute such strength of our approach under zero-shot scenario to two reasons. First, we decompose the relation and separately embed the words corresponding to the subject and object into a semantic space during relation embedding. This is different from modeling the relation holistically using LSTM as in \cite{chen2019weakly}, which lacks flexibility and is hard to learn with limited training data. Second, we treat the relation as a natural language in the reconstruction stage, which enhances the model's ability in visual reasoning through forcing it to infer the remaining words conditioned on the related visual content and the previously generated words in the relation.
\section{Conclusion}
\label{sec:conclusion}
In this paper, we defined a novel task of visual relation grounding in videos which is of significance in underpinning other high-level video-language tasks. To solve the challenges in the task, we proposed a weakly-supervised video relation grounding method by modeling the video as hierarchical spatio-temporal region graph, and collaboratively optimizing two region sequences over it by incorporating relation as textual clues and passing messages by spatial attention shift. Our experiments demonstrated the effectiveness of the proposed approach. Future efforts can either be made on how to jointly ground the subject and object in videos conditioned on their interactions, or how to better capture the temporal dynamics of relations in videos. In addition, it is also important to explore how to better optimize the video graph model based on video-level supervisions only. Another promising direction could be utilizing relation to boost video language grounding and video question answering.
\section*{Acknowledgement}
This research is supported by the National Research Foundation, Singapore under its International Research Centres in Singapore Funding Initiative. Any opinions, findings and conclusions or recommendations expressed in this material are those of the author(s) and do not reflect the views of National Research Foundation, Singapore.


%
%
\bibliographystyle{splncs04}
\bibliography{egbib}
\appendix
\section{Implementation Details}
\textbf{Trajectory Generation}. 
In the test phase, to group continuous frames into candidate segments, two adjacent frames will be grouped as long as the distance between them are smaller than a value of $D_{group}=max(1,10*r)$, where $r$ denotes the frame sample rate of the video, and the maximal sample distance is $\frac{1200}{120}=10$ frames in the dataset. Besides, to obtain the bounding boxes on un-sampled frames, we adopt linear interpolation as follows.
Given two bounding boxes of a trajectory $B_k$ and $B_{k+1}$ corresponding to the sampled frames $F_k$ and $F_{k+1}$, we achieve the bounding boxes in between by
\begin{equation}
    B_c = \frac{F_{k+1}-F_c}{F_{k+1}-F_k}*B_k+\frac{F_c-F_k}{F_{k+1}-F_k}*B_{k+1},~F_c\in(F_k, F_{k+1})
\end{equation}
where $B_c$ denotes the bounding box in frame $F_c$. Since the maximal distance between two sampled frames is 10 (about 1/3 seconds), it is reasonable to think that the objects move linearly in such a short time.

\textbf{Baselines}. 
In both implementation variants of WSSTG \cite{chen2019weakly}, we extract 15 trajectory proposals (resulting in 15*14 = 210 pairs) for each video based on the frame-level region proposals. Each trajectory is then evenly divided into 20 sub-segments, with each segment represented by region appearance feature (average across frames) and sequence feature I3D-RGB as in \cite{chen2019weakly}. Then, the trajectories are modeled with LSTM and interacted with the query sentence to get the similarities between them. 

\section{Additional Results}
\vspace{-5pt}
To evaluate the models' performances on different kinds of relationships, we classify the relationships defined in ImageNet-VidVRD \cite{shang2017video} into dynamic and static ones (Refer to Sec.~\ref{sec:dataset}) and separately report results on them. As a result, we obtain 2343 video-relation instances for static relationships and 2492 for dynamic ones, both covering the 200 test videos. 

As shown in Table \ref{tab:abares}, the results on static relationships are better than those on dynamic ones. When compare our method with the baseline approaches, we can achieve consistently better results as in the main text. We speculate the reason is that moving objects will result in deformation, motion blur and occlusion which are very challenging for grounding. Fortunately, the online optimization mechanism in our approach exploits the relationships to pinpoint some objects with the condition of their related partners, and thus cope better with the dynamic scenario than the baselines. This speculation is also supported by the observation on the co-occurrence baseline that the performance drops significantly from 21.4\% to 13.81\% from static to dynamic scenario, without modeling of the relationships between objects. 

\begin{table*}[t]
\begin{center}
\caption{Grounding results on dynamic and static relationships.}
\label{tab:abares}
\begin{tabular}{l|ccc|ccc|ccc}
\toprule
\multirow{2}{*}{Methods} & \multicolumn{3}{c|}{Dynamic} & \multicolumn{3}{c|}{Static} & \multicolumn{3}{c}{Overall}\\
& $Acc_S$ & $Acc_O$ & $Acc_R$ & $Acc_S$ & $Acc_O$ & $Acc_R$ & $Acc_S$ & $Acc_O$ & $Acc_R$\\
\midrule
\midrule
T-Rank $V_1$ \cite{chen2019weakly}  & 15.19 & 8.27 & 3.40 & 24.05 & 11.36 & 4.96 & 20.27 & 10.68 & 3.99\\
T-Rank $V_2$ \cite{chen2019weakly}  & 15.81 & 5.28 & 1.89 & 23.56 & 7.96 & 4.18 & 20.83 & 7.35 & 3.16\\
Co-occur~~\cite{krishna2018referring}  & \underline{20.05} & \underline{21.21} & \underline{13.81} & \underline{33.43} & \underline{30.82} & \underline{21.4} & \underline{25.90} & \underline{25.23} & \underline{16.48}\\
vRGV (ours) & \textbf{32.47} & \textbf{32.86} & \textbf{22.7} & \textbf{37.53} & \textbf{36.51} & \textbf{26.15} & \textbf{36.77} & \textbf{36.30} & \textbf{24.58}\\
\bottomrule
\end{tabular}
\end{center}
\end{table*}
\begin{table*}[t]
\begin{center}
\caption{Grounding results under different temporal overlap thresholds.}
\label{tab:abatemp}
\begin{tabular}{l|ccc|ccc|ccc}
\toprule
\multirow{2}{*}{Methods} & \multicolumn{3}{c|}{tIoU=0.3} & \multicolumn{3}{c|}{tIoU=0.5} & \multicolumn{3}{c}{tIoU=0.7}\\
& $Acc_S$ & $Acc_O$ & $Acc_R$ & $Acc_S$ & $Acc_O$ & $Acc_R$ & $Acc_S$ & $Acc_O$ & $Acc_R$\\
\midrule
\midrule
T-Rank $V_1$ \cite{chen2019weakly}  & 36.51 & 28.67 & 15.05 & 20.27 & 10.68 & 3.99 & 6.15 & 2.67 & 0.55\\
T-Rank $V_2$ \cite{chen2019weakly}  & \underline{36.99} & 20.70 & 12.81 & 20.83 & 7.35 & 3.16 & 6.19 & 1.30 & 0.21\\
Co-occur~~\cite{krishna2018referring}  & 35.30 & \underline{35.50} & \underline{23.23} & \underline{25.90} & \underline{25.23} & \underline{16.48} & \underline{16.81} & \underline{15.04} & \underline{8.94}\\
vRGV (ours) & \textbf{49.97} & \textbf{48.98} & \textbf{33.16} & \textbf{36.77} & \textbf{36.30} & \textbf{24.58} & \textbf{24.27} & \textbf{22.11} & \textbf{13.69}\\
\bottomrule
\end{tabular}
\end{center}
\end{table*}

\begin{figure*}[t]
\begin{center}
\includegraphics[width=1.0\linewidth]{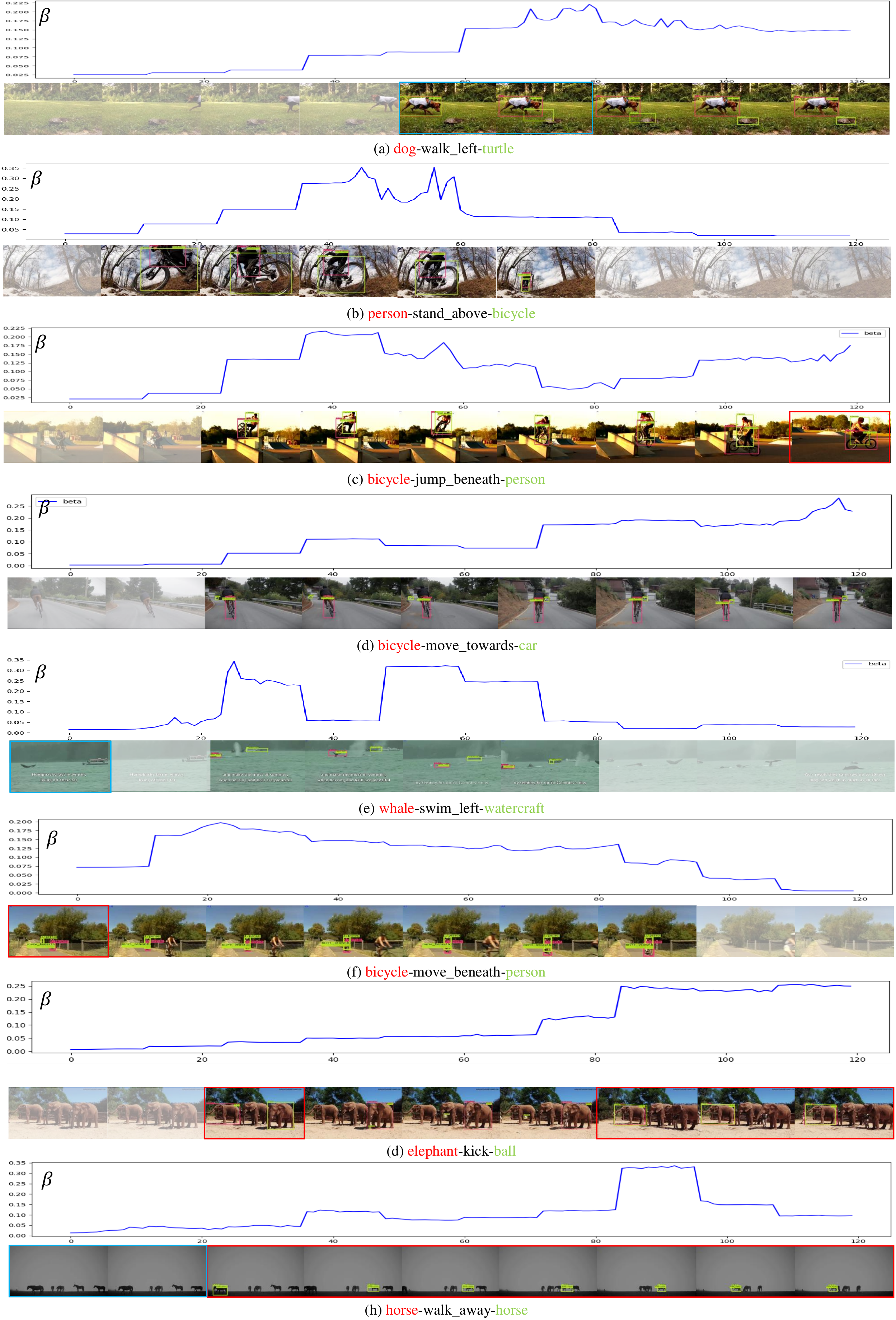}
\end{center}
\caption{Visualization of relation grounding results. The false-positive and false-negative frames are highlighted with red and blue rectangles respectively.}
\label{fig:res}
\end{figure*}

We also compare the models on different temporal overlap thresholds in Table \ref{tab:abatemp}, from which we can draw similar conclusion as in the main text that our method shows superiority to the compared baselines under different settings. As shown in Fig.~\ref{fig:res}, our method can reasonably ground the relation in both space and time. Taking (a) for instance, the learned temporal attention regarding the query relation \emph{dog-walk\_left-turtle} is small when the dog is walking on the right of the turtle, and then it becomes larger when the dog is walking on the left\footnote{According to the dataset definition, the spatial relationships are in camera view.}. Besides, the model can successfully pinpoint the subject \emph{dog} and object \emph{turtle} on the grassland most of the time. Similarly, in (b), our method succeeds in finding the relation \emph{person-stand\_above-bicycle} when the person is standing above the bicycle but not sitting on it after he goes up the slope. 

We specially analyze some failing cases in Fig.~\ref{fig:res}. For the relation \emph{elephant-kick-ball} in (g), our method fails to stop the grounding even though the relation is disappeared, and thus results in false positives. This could be due to that the relation is very transient and further the ball is too small to localize, and hence brings in great challenges in spatio-temporal grounding. In (h), our method wrongly grounds the subject and object on the same visual entity when the subject and object belong to the same category (\eg, horse). We speculate the reason is that we directly take the textual representations of the two words which are indistinguishable in feature space as semantic clues to retrieve the related visual subject and object in the spatial attention unit. Thus, the unit is prone to get a redundant location of the subject as the object. Nevertheless, our model can disambiguate the visual entities of same category in the scenarios where one entity matches the query relation and others do not match. For example, in (c) and (f), there are other people present in the video, but our method can precisely find the person with the bicycle jumps (in (c)) or moves (in (f)) beneath him. Similarly in (d) and (e), there are two cars (d) or two watercrafts (e) present in the videos, and our methods can successfully ground the car/watercraft that satisfy the respective relations. 
\section{Dataset}
\label{sec:dataset}
\vspace{-5pt}
There are 35 objects and 132 predicates (relationships) defined in the ImageNet-VidVRD \cite{shang2017video} dataset. The \textbf{objects} are: \emph{turtle, antelope, bicycle, lion, ball, motorcycle, cattle, airplane, red panda, horse, watercraft, monkey, fox, elephant, bird, sheep, frisbee, giant panda, squirrel, bus, bear, tiger, train, snake, rabbit, whale, sofa, skateboard, dog, domestic cat, person, lizard, hamster, car, zebra}. The predicates can be classified into static relationships and dynamic ones. The \textbf{static relationships} include: \emph{above, beneath, left, right, front, behind, taller, larger, next to, inside, hold, bite, lie above, lie beneath, lie left, lie right, lie inside, lie next to, lie with, stand above, stand beneath, stand left, stand right, stand front, stand behind, stand next to, stand inside,  sit above, sit left, sit right, sit front, sit behind, sit next to, sit inside, stop above, stop beneath, stop left, stop right, stop front, stop behind, stop next to, stop with}. The \textbf{dynamic relationships} include:
\emph{swim behind, walk away, fly behind, creep behind, move left, touch, follow, move away, walk with, move next to, creep above, fall off, run with, swim front, walk next to, kick, creep right, watch, swim with, fly away, creep beneath, run past, jump right, fly toward, creep left, run next to, jump front, jump beneath, past, jump toward, walk beneath, run away, run above, walk right, away, move right, fly right, run front, run toward, jump past, jump above, move with, swim beneath, walk past, run right, creep away, move toward, feed, run left, fly front, walk behind, fly above, fly next to, fight, walk above, jump behind, fly with, jump next to, run behind, move behind, swim right, swim next to, move past, pull, walk left, ride, move beneath, toward, jump left, creep toward, fly left, walk toward, chase, creep next to, fly past, move front, run beneath, creep front, creep past, play, move above, faster, walk front, drive, swim left, jump away, jump with.}
\end{document}